\documentclass{article}
\usepackage{ijcai20}

\usepackage{times}
\usepackage{soul}
\usepackage{url}
\usepackage[hidelinks]{hyperref}
\usepackage[utf8]{inputenc}
\usepackage[small]{caption}
\usepackage{graphicx}
\usepackage{amsmath}
\usepackage{amsthm}
\usepackage{booktabs}
\usepackage{algorithm}
\usepackage{algorithmic}
\usepackage{multirow}
\usepackage{float}
\usepackage{epstopdf}
\usepackage{tabularx}
\usepackage{subfig}
\usepackage{caption}
\usepackage{amsmath}
\usepackage{footmisc}

\title{Unsupervised Vehicle Re-identification with Progressive Adaptation}

\author{
Jinjia Peng$^{1,2}$\and
Yang Wang$^{3,4}$\footnote{Corresponding Author}\and
Huibing Wang$^{1,2}$\and
Zhao Zhang$^{3,4}$\footnotemark[1] \and
Xianping Fu$^{1,2}$\footnotemark[1]\and
Meng Wang$^{3,4}$\\
\affiliations
$^1$College of Information and Science Technology, Dalian Maritime University, Liaoning, Dalian
$^2$Pengcheng Laboratory, Shenzhen, Guangdong \\
$^3$Key Laboratory of Knowledge Engineering with Big Data, Ministry of education, Hefei University of Technology, China \\
$^4$School of Computer Science and Information Engineering, Hefei University of Technology, China \\
\emails
\{jinjiapeng, huibing.wang,fxp\}@dlmu.edu.cn,
yangwang@hfut.edu.cn,
\{cszzhang,eric.mengwang\}@gmail.com
}

\begin{document}

\maketitle

\begin{abstract}
  Vehicle re-identification (reID) aims at identifying vehicles across different non-overlapping cameras views. The existing methods heavily relied on well-labeled datasets for ideal performance, which inevitably causes fateful drop due to the severe domain bias between the training domain and the real-world scenes; worse still, these approaches required full annotations, which is labor-consuming. To tackle these challenges, we propose a novel progressive adaptation learning method for vehicle reID, named PAL, which infers from the abundant data without annotations. For PAL, a data adaptation module is employed for source domain, which generates the images with similar data distribution to unlabeled target domain as ``pseudo target samples''. These pseudo samples are combined with the unlabeled samples that are selected by a dynamic sampling strategy to make training faster. We further proposed a weighted label smoothing (WLS) loss, which considers the similarity between samples with different clusters to balance the confidence of pseudo labels. Comprehensive experimental results validate the advantages of PAL on both VehicleID and VeRi-776 dataset.
\end{abstract}

\section{Introduction}
\noindent Vehicle-related research has attracted a great deal of attention, ranging from vehicle detection, tracking \cite{tang2019cityflow} to classification \cite{ref_article5}. Unlike them, vehicle re-identification (reID) aims to match a specific vehicle across scenes captured from multiple non-overlapping cameras, which is of vital significance to intelligent transport. Most of the existing vehicle reID methods, in particular for deep learning models, usually adopt the supervised approaches \cite{zhao2019structural,lou2019embedding,bai2018group,wang2017orientation,guo2019two} for an ideal performance. However, they suffer from the following limitations.

\emph{On one hand}, due to the domain bias, well-trained vehicle reID models under these supervised methods may suffer from a poor performance when directly deployed to the real-world large-scale camera networks. \emph{On the other hand}, these methods heavily relied on the full annotations, i.e., the identity labels of all the training data from multiple cross-view cameras. However, it is labor expensive to annotate large-scale unlabeled data in the real-world scenes. In particular for the vehicle reID task, it is always required to annotate the same vehicle under all cameras. Hence, how to incrementally optimize the vehicle reID algorithms utilizing the combination of the abundant unlabeled data and existing well-labeled data is practical but challenging.

\begin{figure*}[ht]
\centering
\includegraphics[width=15cm]{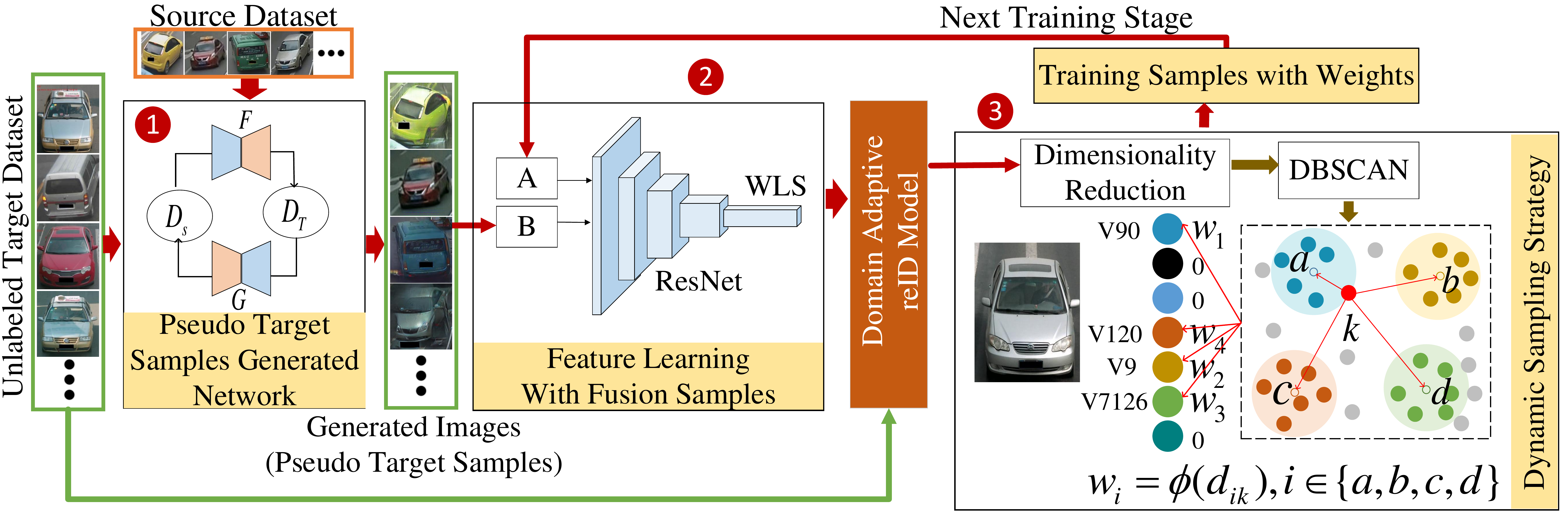}
\caption{Illustration of the PAL framework. The images are transferred from the source domain to target domain for pseudo target samples generation. During each iteration, we 1) train the reID model by the proposed WLS-based feature learning network that utilizes the fusion data, combining the pseudo target samples with selected samples, and 2) assign pseudo labels with various weights for unlabeled images and select reliable samples according to a dynamic sampling strategy. A means pseudo-labeled images and B is generated images.} \label{fig1}
\end{figure*}

To these ends, a few unsupervised strategies have been proposed. Specifically, \cite{bashir2019vr} takes the cluster to be pseudo labels and then select the reliable pseudo-labeled samples for training. However, the incorrect annotations assigned by clustering are inevitable.
One may try transferring images from the well-labeled domain to the unlabeled domain via style transfer for the unsupervised vehicle reID \cite{isola2017image,yi2017dualgan}. The generated images are employed to train the reID model, which preserves the identity information from well-labeled domain, while learn the style of unlabeled domain. However, this solution is limited by the learned style that is different from the unlabeled domain, and may fail to adapt to the real-world scenes without label information.

In this paper, we propose a novel unsupervised method, named PAL, together with Weighted Label Smoothing (WLS) loss to better exploit the unlabeled data, while adapt the target domain to vehicle reID ``\emph{progressively}''. Unlike the existing unsupervised reID methods, a novel adaptation module is proposed to generate ``pseudo target images'', which learns the style of unlabeled domain and preserves the identity information of the labeled domain. Besides, dynamic sampling strategy is proposed to select reliable pseudo-labeled data from the clustering result. Furthermore, the fusion data that combines the ``pseudo target images'' and reliable pseudo-labeled data is employed to train the reID model in the subsequent training. To facilitate the presentation, we illustrate the major framework in Fig. \ref{fig1}.

Our major contributions are summarized as follows:
\begin{itemize}
\item A novel progressive method, named PAL, is proposed for unsupervised vehicle reID to better adapt the unlabeled domain, which iteratively updates the model by WLS based feature learning network, and adopts dynamic sampling strategy to assign labels for selected reliable unlabeled data.
\item To make full use of the existing labeled data, PAL employs a data adaptation module based on Generative Adversarial Network (GAN) for generating images as the ``pseudo target samples'', which are combined with the selected samples from unlabeled domain for model training.
\item The feature learning network with WLS loss is proposed, which considers the similarity between the samples and different clusters to balance the confidence of pseudo labels to improve the performance.
\end{itemize}
Experimental results on benchmark data sets validate the superiority of our method, which is even better than some supervised vehicle reID approaches.

\section{Progressive Adaptation Learning for Vehicle ReID}
In this section, we formally discuss our proposed technique of progressive adaption learning, named PAL, for vehicle reID. Specifically, as shown in Fig.\ref{fig1}, a data adaptation module based on GAN is trained to transfer the well-labeled images to the unlabeled target domain, which aims at smoothing the domain bias and make full use of the existing source domain images. Then the generated images are employed as the ``pseudo target samples'' and combined with selected unlabeled samples to serve as the input to ResNet50 \cite{He2015} for feature learning, which adapts the target domain progressively. When the model is trained, WLS loss is proposed to balance the confidence of unlabeled samples and different clusters, which exploits the pseudo labels with different weights, according to the model trained by the last iteration. Then the output features of the reID model are employed to select reliable samples by dynamic sampling strategy. Finally, the ``pseudo target samples'' with accurate labels and selected samples from unlabeled domain with pseudo labels are combined to be the training sets for the next iteration. In this way, a more stable adaptive model could be learned progressively.

\subsection{Pseudo Target Samples Generated Network}
For a target domain, the supervised learning approaches are limited by the unlabeled samples, which can't be utilized to train reID model. Though there are well-labeled datasets, directly applying them to target domain may suffer from a poor performance because of the domain bias mainly caused by diversified illuminations and complicated environment. To remedy this, CycleGAN \cite{zhu2017unpaired,wu2019cycle} is employed to make full use of these well-labeled data, which generates \textbf{``pseudo target samples''} to narrow down the domain bias by transferring the style between source domain and target domain. The generated images share the similar style with the target domain while preserving the identity information of the source domain.
Specifically, it comprises of two generator-discriminator pairs, ($G$, $D_{T}$) and ($F$, $D_{S}$), which maps a sample from source (target) domain to target (source) domain and generate a sample which is indistinguishable from those in the target(source) domain\cite{almahairi2018augmented}. For PAL, besides the traditional adversarial losses and cycle-consistent loss, a content loss \cite{ref_article27} is utilized to preserve the label information from the source domain, which is formulated as:
\begin{equation}
\begin{split}
L_{id}(G,F,X,Y) &= E_{y\sim p_{data}(y)}||F(y)- y||_1 \\ & +E_{x\sim p_{data}(x)}||G(x)- x||_1,
\end{split}
\end{equation}
where $X$ and $Y$ represent the source domain and target domain, respectively. $p_{data}(x)$ and $p_{data}(y)$ represent the sample distributions in the source and target domain.

One may wonder why the generated network could make full use of the well-labeled data, we answer this question from the following two aspects:
\begin{itemize}
\item Through CycleGAN, the generated ``pseudo target samples'' have similar distribution for the target domain, which reduces the bias between source and target domain.
\item Furthermore, the identity information of source domain is also preserved by turning the content loss during the transferring phrase, implying that the well labeled annotations could be re-utilized in the subsequence.
\end{itemize}
\subsection{Feature Learning Network with WLS Loss}
Feature learning plays a vital role of the PAL, which trains the model by combining the generated ``pseudo target images'' with the selected pseudo labeled samples. For the ``pseudo target images'', it's easy to obtain the label information. However, how to assign labels for the pseudo labeled samples reasonably is a big challenge, due to the following facts:
\begin{itemize}
\item If the clustering centroids serve as the pseudo labels, it may cause ambiguous prediction during the training phase due to the inaccurate clustering results.
\item Moreover, it is not reasonable to assign the same labels to all the samples regardless of the distance to the clustering centroids.
\end{itemize}
Hence, WLS loss is proposed to set the pseudo label distribution as a weighted distribution over all clusters, which effectively regularizes the feature learning network to the target training data distribution.

Specifically, we model the virtual label distribution as a weighted distribution over all clusters for unlabeled data according to the distance between the features and each centroid of clusters. Thus, the weights over all clusters are different in WLS loss. In this way, a dictionary $\alpha$ is constructed to record the weights. For an image $g$, the weights of the label can be calculated as:
\begin{equation}
w^g_k = \frac{1}{K}{\alpha^g _k}, k\in[1,K],
\end{equation}
where $\alpha^g _k$ represents the weight of the image $g$ over the $k$-th cluster. To obtain $\alpha^g_k$, unlabeled samples are clustered to obtain centroids set $C=\{c_1,c_2,...,c_k\}, k\in[1,K]$, which is introduced in section \ref{sec:DSS}. $K$ is the number of clusters, while the similarity between $g$ and $c_k$ can be calculated as $d_k^g=||f_g-f_{c_k}||_2$, where $f$ represents the feature of images or centroids. The set of distance of image $g$ over $K$ centroids could be described as: $d^g=\{d^g_1,d^g_2,...,d^g_k\},  k\in[1,K]$. Inspired with \cite{huang2019multi-pseudo}, then all elements in $d^g$ are sorted with descending order, and saved to $ds^g$. $\alpha _k$ is obtained by taking the corresponding index of $ds^g_k$ in the set of $ds^g$:
\begin{equation}
\alpha _k^g= (1- \frac{d^g_k}{max(d^g)}) \cdot  \psi _{ds^g}(d^g_k),
\end{equation}
where $\psi _{ds^g}(\cdot)$ is the index of $d^g_k$ in $ds^g$. Thus, the corresponding relationship between images and cluster centroids is constructed with different weights. In order to filter noise, top-$m$ in $w$ are selected as reliable weights, with others set to 0. To this end, we have:
\begin{equation}
  w _k^g=
  \begin{cases}
    0, &\text{if $w _k^g < \theta $}\\
	w _k^g, &\text{if $w _k^g >= \theta$,}
  \end{cases}
\end{equation}
where $\theta$ is a threshold. Thus, the WLS loss of unlabeled data $\ell_{wls}$ can be formulated as:
\begin{equation}
\ell_{wls}= - \sum_{k=1}^{K}w_k{log(p(k))}
\end{equation}
Besides the real unlabeled samples, there are some generated images by CycleGAN that are combined to train the reID model. The training loss is defined as follows:
\begin{equation}
\begin{split}
\ell &= - (1-\sigma) \cdot {log(p(y))} - \sigma \cdot \lambda \cdot \sum_{k=1}^{K}w_k{log(p(k))}
\end{split}
\end{equation}
For a generated image $\lambda=0$, the loss is equivalent to cross-entropy loss and $y$ is the label of the generated image. When $\lambda=1$ means the image is from the unlabeled data and $y$ is the cluster it belongs to. Beyond that, for the unlabeled data, $\sigma$ is a smoothing factor between cross-entropy loss and WLS loss.

\subsection{Dynamic Sampling Strategy}\label{sec:DSS}
It is crucial to obtain appropriately selected candidates to exploit the unlabeled data. When the model is weak, small reliable measure is set, which is nearby to the cluster centroids in the feature space. As the model becomes stronger in subsequent iterations, various instances should be adaptively selected as the training examples. Hence, a dynamic sampling strategy is proposed to ensure the reliability of selected pseudo-labeled samples. As shown in Fig.\ref{fig1}, images in the target domain is processed by the well-trained reID model to output the features with high dimensions. Most methods select the K-Means to generate clusters, which are required to be initialized by the cluster centroids. However, it is uncertain on how many categories are required in the target domain. Hence, DBSCAN is selected as the clustering method. Specifically, instead of employing the fixed clustering radius, the paper employs a dynamic cluster radius $rad$ that is calculated by K-Nearest Neighbors (KNN). After DBSCAN, in order to filter noise, some of top reliable samples are selected to be assigned with soft labels, according the distance between features of the samples and cluster centroids. For our method, samples with $||f_g,c_{f_g}||_2< \gamma$ are satisfied for the next iteration for training model, where $f_g$ is the feature of $g$-th image and $c_{f_g}$ is the feature of the cluster centroid where the $f_g$ belongs to. Our method is summarized in Algorithm \ref{alg_1}.

\begin{algorithm}[ht]
\caption{PAL for Unsupervised Vehicle reID}
\begin{algorithmic}[1]
\REQUIRE
 Numbers of images on the target domain $N$,
 labeled Source domain $S$,
 unlabeled target domain $T$,
 iteration number $M$,
 cluster number $K$,
 reliability threshold $\gamma$,
 new Training Set $D$
\ENSURE
 An encoder $E$ for target domain
\STATE Transfer style from $S$ to $T$ by GAN to generate pseudo target images $ST$
\STATE Initialization $E^{(0)}$ with $ST$, $D$: $D = ST$
\FOR{ $i$=1 to $M$}
\STATE   Train $E^{(i)}$ with $D$ utilizing WLS-based feature learning network, compute $ft=E^{(i)}(D)$
\STATE   Reduce dimension by manifold learning $f=mad(ft)$, calculate number and centroids of clusters: ($K, C)=DBSCAN(f)$
\STATE   Select features of centroids: ${\{c_k\}^{K}_{k=1}}\to{\{f_{c_k}\}^{K}_{k=1}}$
\STATE   $D$= $ST$
\FOR{$k$=1 to $K$}
\FOR{$g=1$ to $N$}
\IF {$||f_g,f_{c_k}||_2>\gamma$}
\STATE      {$D = D \cup T_g$}
\STATE      {Calculate weights $w_k^g$ by Eq.(2)(3)(4)}
\ENDIF
\ENDFOR
\ENDFOR
\ENDFOR
\end{algorithmic}
\label{alg_1}
\end{algorithm}

\section{Experiments}
\subsection{Datasets and Evaluation Metrics}
The experiments are conducted over the following two typical data sets for Vehicle reID: VeRi-776 and VehicleID. \textbf{VeRi-776} \cite{ref_article20} is a large-scale urban surveillance vehicle dataset for reID, which contains over 50,000 images of 776 vehicles, where 37,781 images of 576 vehicles are employed as training set, while 11,579 images of 200 vehicles are employed as a test set. A subset of 1,678 images in the test set generates the query set. \textbf{VehicleID} \cite{ref_article11} is a surveillance dataset from the real-world scenario, which contains 221,763 images corresponding to 26,267 vehicles in total. From the original testing data, four subsets, which contain 800, 1,600, 2,400 and 3,200 vehicles, are extracted for vehicle search for multi-scales.
CMC curve and mAP are employed to evaluate the overall performance for all test images. For each query, its average precision (AP) is computed from the precision-recall curve.

\subsection{Implementation Details}
For CycleGAN, the model is trained in the tensorflow \cite{abadi2016tensorflow}. It is worth mentioning that any label notation aren't utilized during the learning procedure. In the stage of the feature learning, the ResNet50 \cite{He2015} is employed as the backbone network. For PAL, the images are transferred by CycleGAN from source domain to target domain, which are as ``Pseudo target samples'' for training the feature learning model. Considering the limit of device, when training reID model on the VeRi-776, 10,000 transferred images from VehicleID are utilized as the ``pseudo target images''. The same implementations are conducted when the reID model is trained on the VehicleID. Besides that, when training the unsupervised model on VehicleID, only 35,000 images from the VehicleID are selected as the training set. Moreover, any annotations of target domain aren't employed in our framework.

\begin{figure}[htbp]
\centering
\includegraphics[width=5cm]{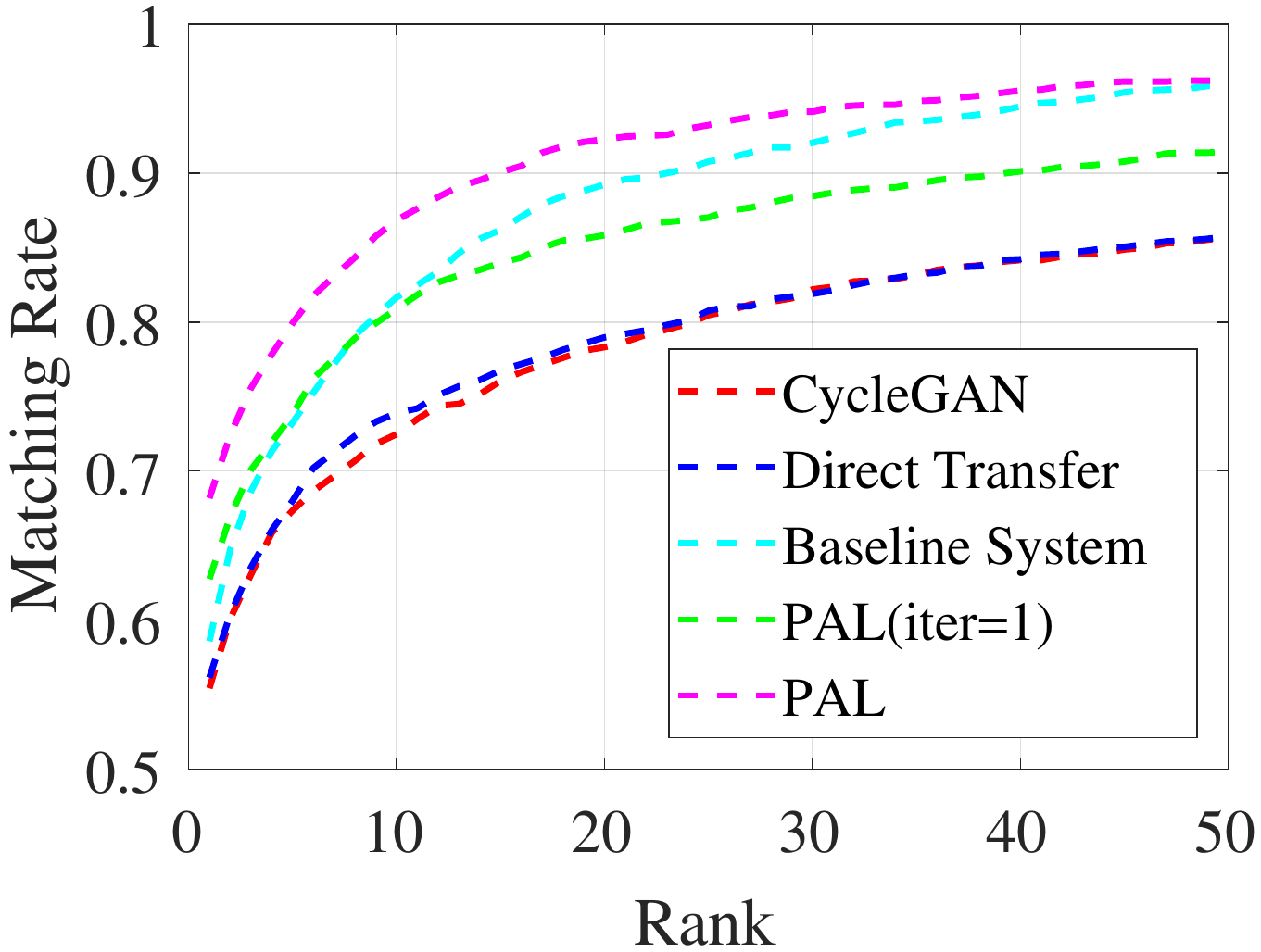}
\caption{ CMC results of several typical methods on VeRi-776. The proposed PAL outperforms other compared methods, especially the ``CycleGAN'' and the ``Direct Transfer''. Besides that, compared with the ``PAL (iter=1)'', ``PAL'' has a high improvement in mAP, which demonstrates the progressive learning could increase the adaptive ability for the reID model in the unlabeled target domain.} \label{R_VeRi_C}
\end{figure}

\begin{table}[ht]
\small
\centering
\begin{tabular}{p{3.2cm}|p{1.1cm}|p{1.1cm}|p{1.1cm}}
\hline
 Method &  mAP(\%) & Rank1(\%) & Rank5(\%)\\
\hline
\hline
FACT     &18.75 &52.21 &72.88 \\
FACT+Plate-SNN+STR &  27.77 & 61.44 & 78.78\\
\hline
VR-PROUD & 22.75 & 55.78 & 70.02 \\
PUL & 17.06 & 55.24 & 66.27\\
CycleGAN & 21.82 & 55.42 & 67.34\\
\hline
Direct Transfer & 19.39 & 56.14 & 68.00\\
Baseline System & 31.94 & 58.58 & 73.24\\
\hline
PAL & \textbf{42.04}& \textbf{68.17}& \textbf{79.91}\\
\hline
\end{tabular}
\caption{Performance of different methods on VeRi-776.  The best
results are shown in bold face. PAL can achieve best performance.}\label{VeRi-STA}
\end{table}

\begin{table*}[htbp]
\footnotesize
\centering
\begin{tabular}
{p{2.2cm}|p{0.8cm}|p{0.8cm}|p{0.8cm}|p{0.8cm}|p{0.8cm}|p{0.8cm}|p{0.8cm}|p{0.8cm}|p{0.8cm}|p{0.8cm}|p{0.8cm}|p{0.8cm}}
\hline
\multirow{2}*{Method} & \multicolumn{3}{c|}{Test size = 800 (\%)} & \multicolumn{3}{c|}{Test size = 1600 (\%)} & \multicolumn{3}{c|}{Test size = 2400 (\%)} & \multicolumn{3}{c}{Test size = 3200 (\%)}\\
\cline{2-13}              & mAP    & Rank1 &Rank5  &mAP    &Rank1  &Rank5 & mAP    & Rank1 &Rank5  &mAP    &Rank1  &Rank5\\
\hline
\hline
 FACT                    & --  &49.53  &67.96  &--  &\textbf{44.63}  &64.19 &--  &39.91  &60.49  &--  &--  &--\\
 Mixed Diff+CCL          & --  &49.00  &\textbf{73.50}  &--  &42.80  &\textbf{66.80} &--  &38.20  &\textbf{61.60}  &--  &--  &--\\
\hline
 PUL                     & 43.90  &40.03  &56.03  &37.68  &33.83  &49.72 &34.71  &30.90 &47.18 &32.44 & 28.86 & 43.41\\
 CycleGAN                & 42.32  &37.29  &58.56  &34.92  &30.00  &49.96 & 31.89  &27.15  &46.52  &29.17  &24.83  &42.17\\
\hline
 Direct Transfer         & 40.58  &35.48  &57.26  &33.59  &28.86  &48.34 & 30.50  &26.08  &44.02  &27.90  &23.85  &39.76\\
 Baseline System         & 42.96  &39.11  &55.24  &38.03  &34.04  &50.91 & 34.04  &30.10  &48.41  &31.98  &28.24  &43.77\\
\hline
 PAL                & \textbf{53.50}  &\textbf{50.25}  &64.91  &\textbf{48.05}  &44.25  &60.95 & \textbf{45.14}  &\textbf{41.08}  &59.12  &\textbf{42.13}  &\textbf{38.19}  &\textbf{55.32} \\
\hline
\end{tabular}
\caption{Performance of various methods over different reID methods on VehicleID.  It is notable that the best results are shown in bold face.
PAL can achieve best performance in most situations. Mixed Diff+CCL can also achieve good performance.}\label{VehicleID-STA}
\end{table*}

\begin{table}[ht]
\small
\centering
\begin{tabular}{p{1.4cm}|p{1.4cm}|p{1.4cm}|p{1.0cm}|p{1.0cm}}
\hline
 Method &  Generated Images & Original Images & WLS & CE\\
\hline
BS & $\times$ & $\surd$ & $\times$ & $\surd$ \\
CEL & $\surd$ & $\times$ & $\times$ & $\surd$ \\
OIMG & $\times$ & $\surd$ & $\surd$ &$\times$ \\
PAL & $\surd$ & $\times$ & $\surd$ & $\times$ \\
\hline
\end{tabular}
\caption{The settings for different ablation models.}\label{tabModels}
\end{table}

\subsection{Comparison with the State-of-the-art Methods}
\begin{figure}[ht]
\centerline{
\subfloat[Test size=800]{\includegraphics[width=1.6in,height=1.3in]{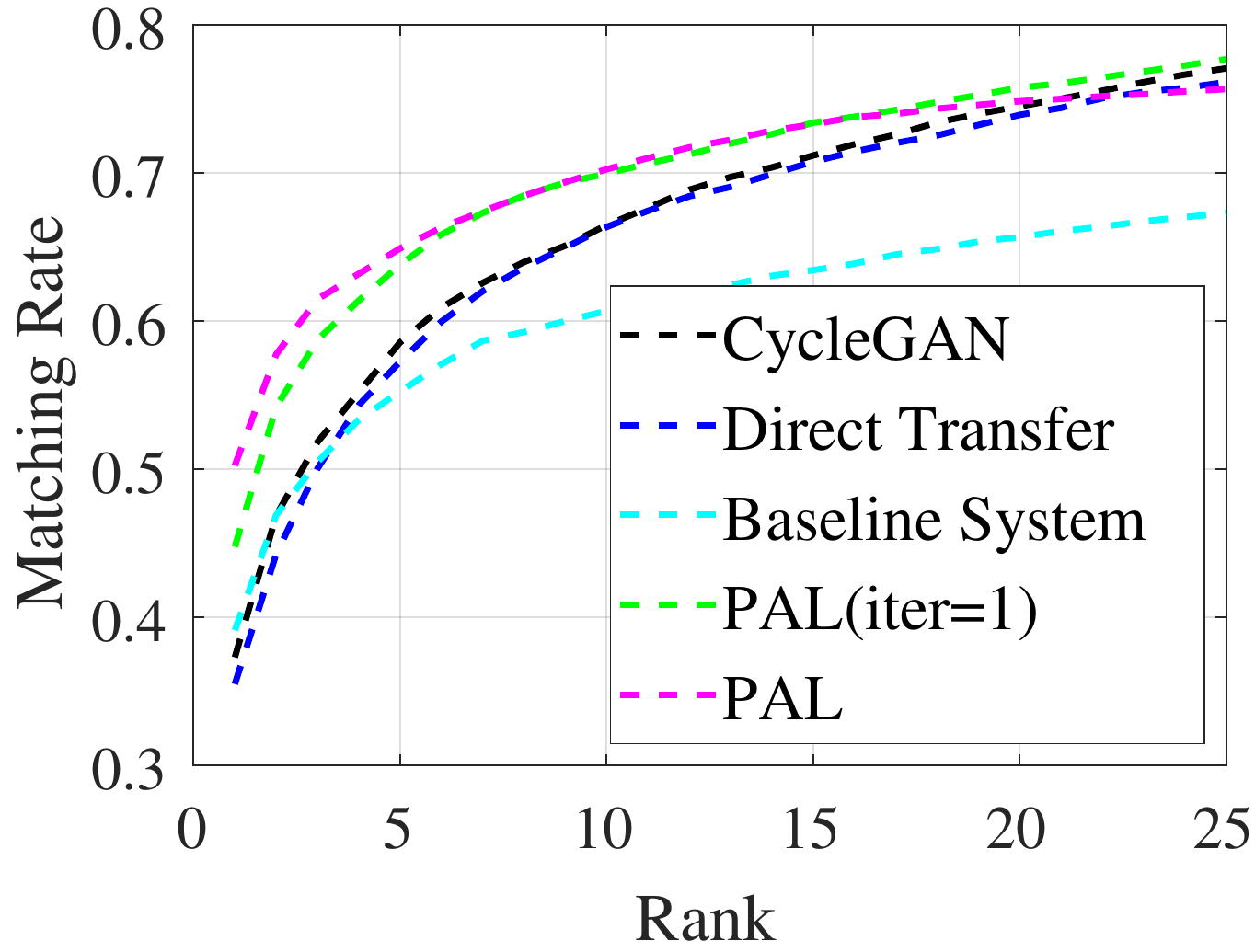}}
\subfloat[Test size=1600]{\includegraphics[width=1.6in,height=1.3in]{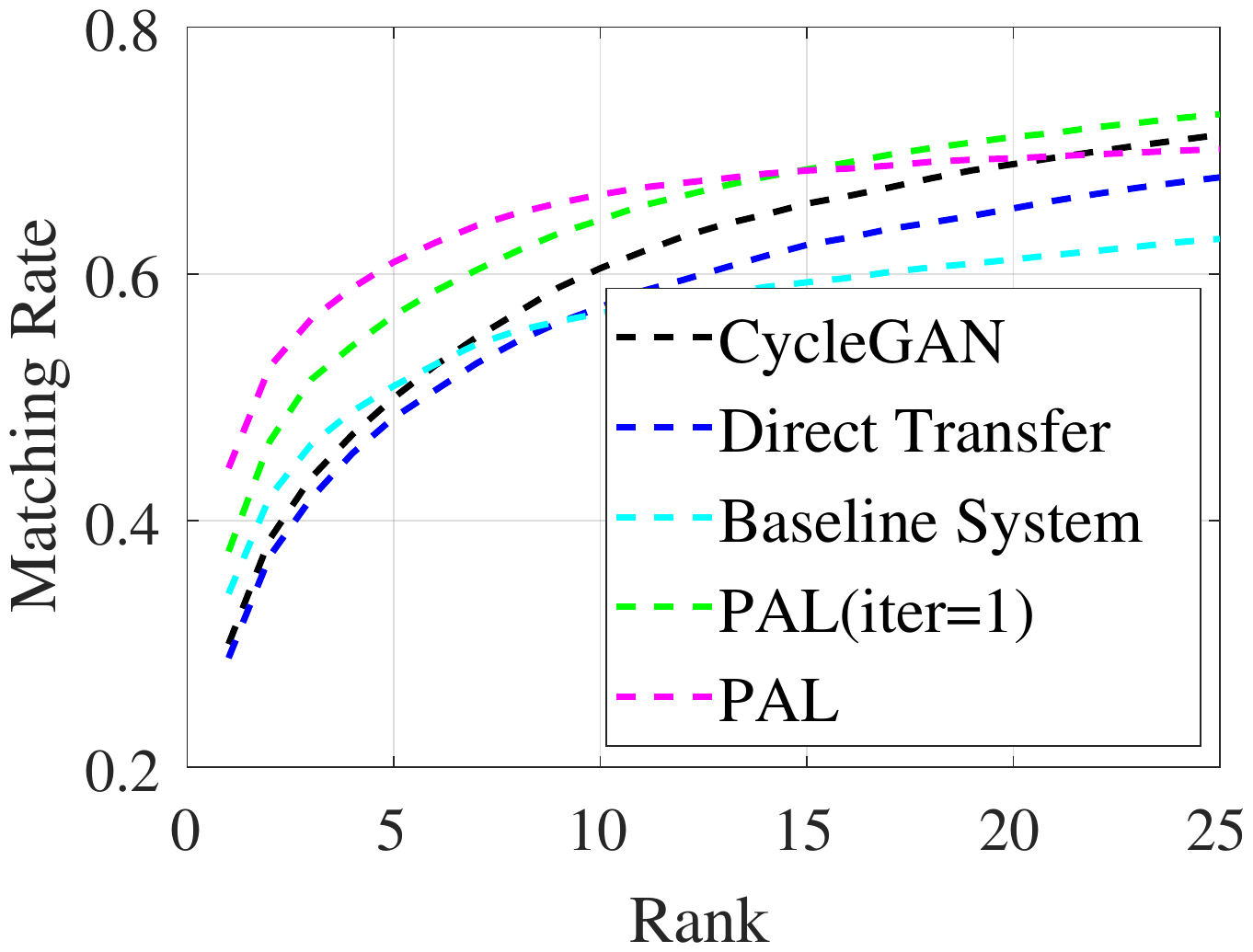}}
}
\centerline{
\subfloat[Test size=2400]{\includegraphics[width=1.6in,height=1.3in]{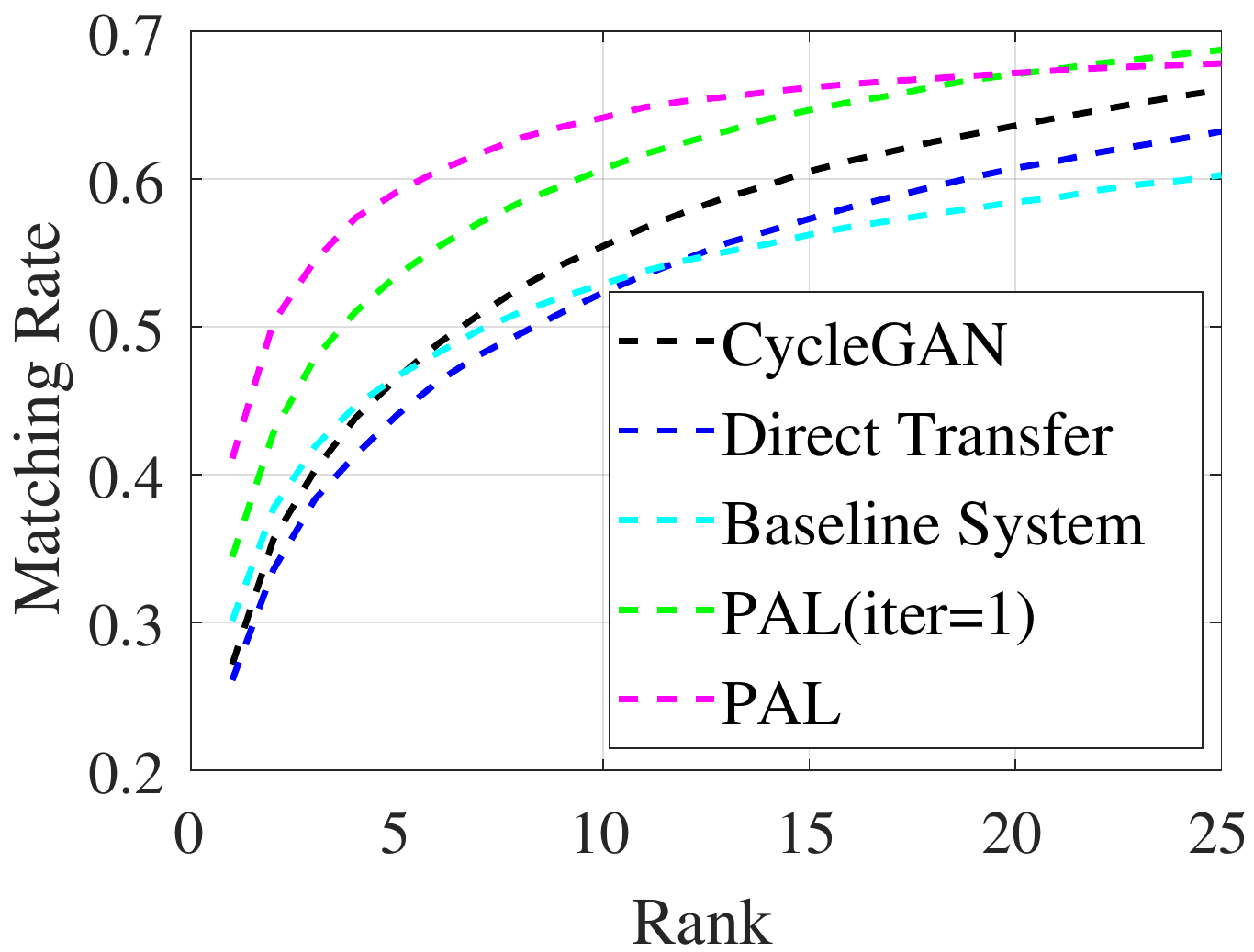}}
\subfloat[Test size=3200]{\includegraphics[width=1.6in,height=1.3in]{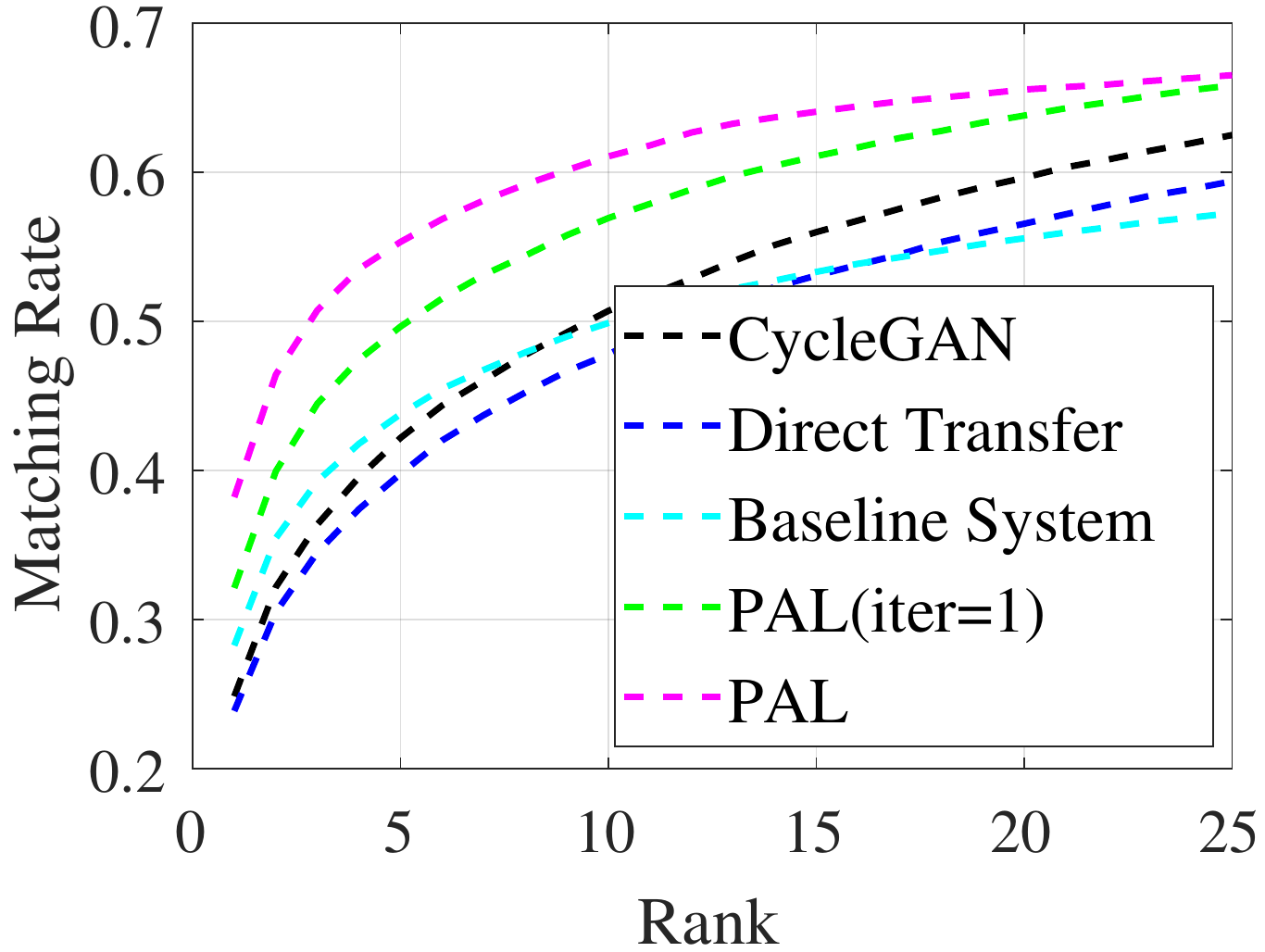}}
}
\caption{CMC curves of several typical methods on VehicleID. From the curves, it is obvious that better results could be achieved when the model is tested on the different datasets, which demonstrates PAL is effective for different test sets.} \label{R_VehicleID_CMC}

\end{figure}

In this section, the results of the comparison between PAL and other state-of-the-art methods are reported in Tables \ref{VeRi-STA}, \ref{VehicleID-STA} and Figures \ref{R_VeRi_C}, \ref{R_VehicleID_CMC}, which includes: (1) FACT \cite{liu2016deep}; (2) FACT+Plate-SNN+STR \cite{liu2016deep}; (3) Mixed Diff+CCL \cite{ref_article11}; (4)VR-PROUD \cite{bashir2019vr}; (5) CycleGAN \cite{zhu2017unpaired}. This is method of style transfer, which is employed for the domain adaptation; (6) Direct Transfer: It directly employed the well-trained reID model by the \cite{ref_article32} on source domain to the target domain; 7)Baseline System. Compared with the framework of PAL, it utilizes the original samples from source domain instead of generated data and the reID model is only trained with cross-entropy (CE) loss; 8)PUL \cite{fan2018unsupervised}. The methods of (1), (2) and (3) are supervised vehicle reID approaches. And others are unsupervised methods. Specially, the PUL is an unsupervised adaptation method of person reID. Since only a few works focused on the unsupervised vehicle reID, PUL is compared with the proposed PAL in this paper. There are some other methods that similar with PUL. However, most of them require special annotations, such as labels for segmenting or detecting keypoints, which are not annotated in the existing vehicle reID datasets. From the Tables \ref{VeRi-STA}, \ref{VehicleID-STA}, we note that the proposed method achieves the best performance among the compared with methods with Rank-1 = 68.17\%, mAP = 42.04\% on VeRi-776, Rank-1 = 50.25\%, 44.25\%, 41.08\%, 38.19\%, mAP = 53.50\%, 48.05\%, 45.14\%, 42.13\% on VehicleID with the test size of 800, 1600, 2400, 3200, respectively.

Compared with PUL \cite{fan2018unsupervised} and VR-PROUD \cite{bashir2019vr}, PAL has 24.98\% and 19.29\% gains on VeRi-776, respectively.  Our model also outperforms the PUL and VR-PROUD in Rank-1, Rank-5 and mAP on VehicleID. For these methods, the K-Means is employed to assign pseudo-labels for unlabeled samples. Due to the uncertainty on how many categories, the K-Means is not appropriate to be utilized in the reID task. In addition, compared with ``Direct Transfer'', it is obvious that our proposed PAL achieves 22.65\% and 12.03\% gains in mAP and Rank-1 on VeRi-776. It also has similar improvements on VehicleID.  Furthermore, compared with the supervised approaches, such as FACT \cite{liu2016deep}, Mixed Diff+CCL \cite{ref_article11} and FACT+Plate-SNN+STR \cite{liu2016deep}, PAL achieves improvements on VeRi-776 and VehicleID, validating that PAL is more adaptive to different domains.

Compared with the CycleGAN \cite{zhu2017unpaired} that adapts the domain bias by style transfer, our method has large improvements on both VeRi-776 and VehicleID. The proposed PAL achieves 20.22\% and 12.75\% improvements in mAP and Rank-1 on VeRi-776, respectively. Similarly, our method has 12.96\%, 14.25\%, 13.93\% and 13.36\% gains in Rank-1 on VehicleID with the test sets of 800, 1600, 2400 and 3200. The significant improvements are mainly due to the fact that PAL exploits the similarity among unlabeled samples through iteration for unsupervised vehicle reID. Though the generated images have the style of target domain, they are just served as the pseudo samples.  The real samples in the target domain could be more reliable to generate the discriminative features during the stage of training. These results suggest that reliable samples in target domain is an important component for the unsupervised reID task, which indicate that PAL could make full use of the unlabeled samples in target domain.

\begin{table}[ht]
\footnotesize
\centering
\begin{tabular}{p{0.9cm}|p{0.7cm}|p{0.7cm}|p{0.7cm}|p{0.7cm}|p{0.7cm}|p{0.7cm}}
\hline
\multirow{2}*{Iteration} & \multicolumn{3}{c|}{CEL (\%)} & \multicolumn{3}{c}{BS (\%)} \\
\cline{2-7} & mAP & Rank1 & Rank5  & mAP & Rank1 & Rank5 \\
\hline
iter1 &  27.71&  59.89 & 72.10   & 25.04& 57.33& 71.33\\
iter2 &  33.76& 64.12 & 78.06    & 30.19& 58.40& 73.53\\
iter3 &  35.73& 65.55 & 78.18    & 32.49& 59.41& 73.06\\
iter4 &  36.01& 63.28 & 77.47    & 32.63& 59.77& 74.07\\
iter5 &  33.86& 60.90 & 77.11    & 32.86& 60.96& 74.91\\
iter6 &  34.03& 62.09 & 75.38    & 31.94& 58.58& 73.24\\

\hline
\end{tabular}
\caption{Performance of comparison between CEL and BS on VeRi-776.}\label{VeRi-CE}
\end{table}

\begin{table}[ht]
\footnotesize
\centering
\begin{tabular}{p{0.9cm}|p{0.7cm}|p{0.7cm}|p{0.7cm}|p{0.7cm}|p{0.7cm}|p{0.7cm}}
\hline
\multirow{2}*{Iteration} & \multicolumn{3}{c|}{CEL (\%)} & \multicolumn{3}{c}{BS (\%)} \\
\cline{2-7} & mAP & Rank1 & Rank5   & mAP & Rank1 & Rank5 \\
\hline
iter1 &  35.69& 31.43 & 49.54  & 34.93& 30.71& 48.41\\
iter2 &  35.89& 31.44 & 49.65  & 34.00& 29.90& 46.98\\
iter3 &  36.49& 32.39 & 50.84  & 34.33& 30.29& 47.14\\
iter4 &  36.62& 32.73 & 51.01  & 33.62& 29.67& 46.38\\
iter5 &  37.33& 33.69 & 51.25  & 34.08& 30.15& 46.94\\
iter6 &  37.12& 33.45 & 51.36  & 34.04& 30.10& 46.94\\
\hline
\end{tabular}
\caption{Performance of comparison between CEL and BS on VehicleID(2400).}\label{VehicleID-CE}
\end{table}

Compared with ``Baseline System'', PAL has large improvements both on VeRi-776 and VehicleID. The PAL achieves 10.1\% increase in mAP on VeRi-776, and 10.54\%, 10.02\%, 11.1\%, 10.15\% improvements in mAP on VehicleID with different test sets, respectively. These indicate that the ``pseudo target images'' and ``weighted label smoothing'' are two core components in PAL which lead the reID model trained by our method to be more robust to different domains. We discuss more details in the next section.

\subsection{Ablation Studies}
\begin{figure}[ht]
\centerline{
\subfloat[mAP]{\includegraphics[width=1.05in,height=1in]{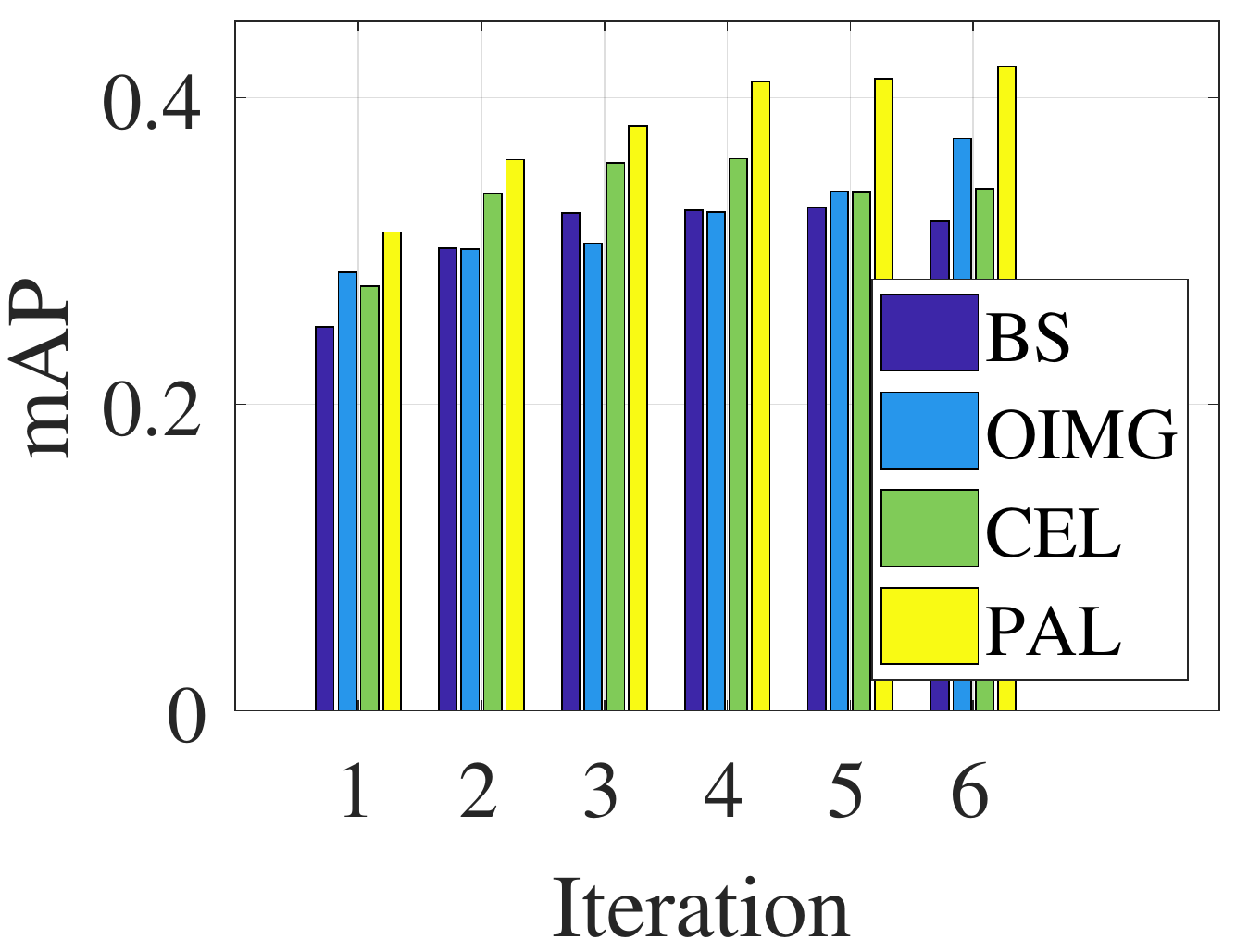}}
\subfloat[Rank-1]{\includegraphics[width=1.05in,height=1in]{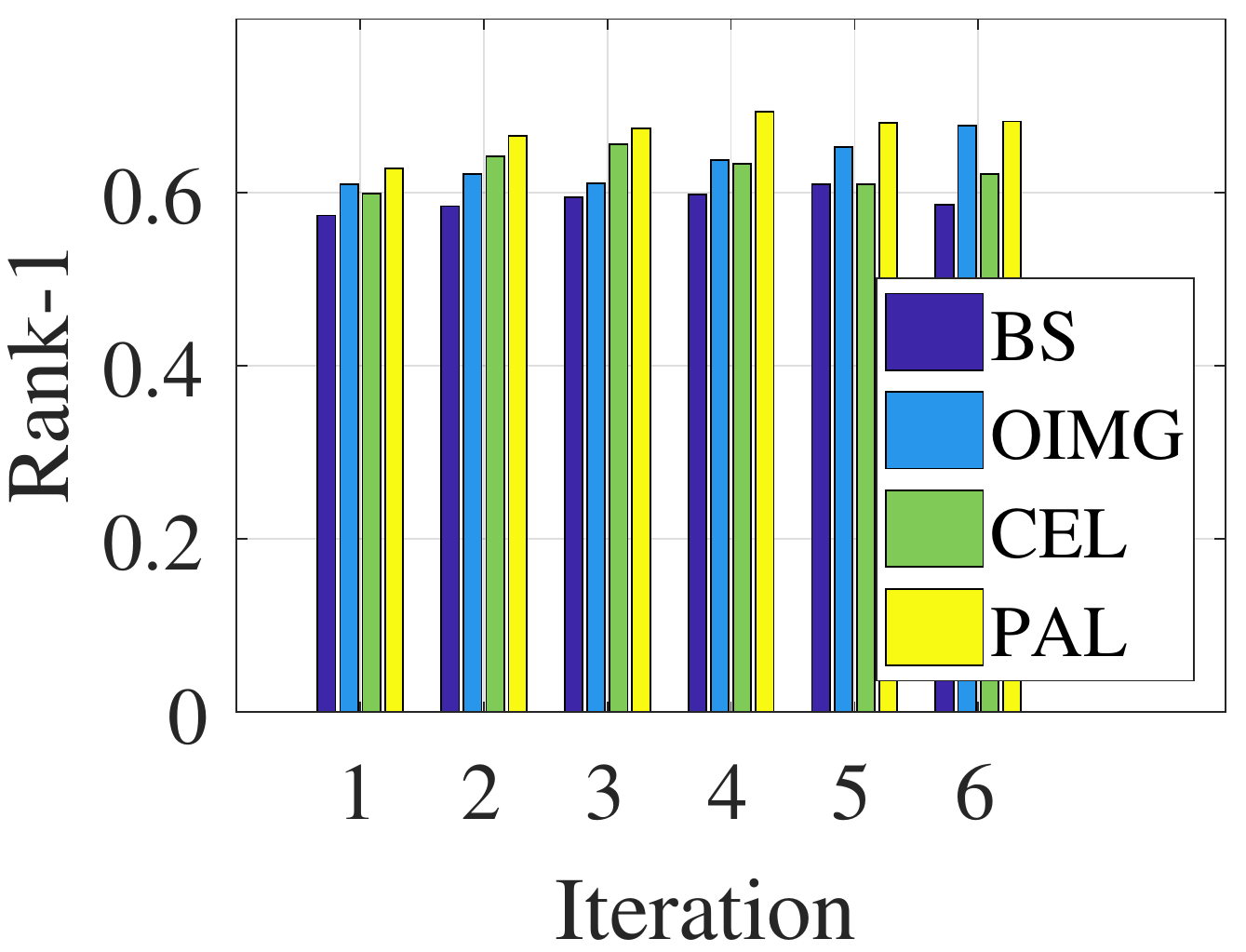}}
\subfloat[Rank-5]{\includegraphics[width=1.05in,height=1in]{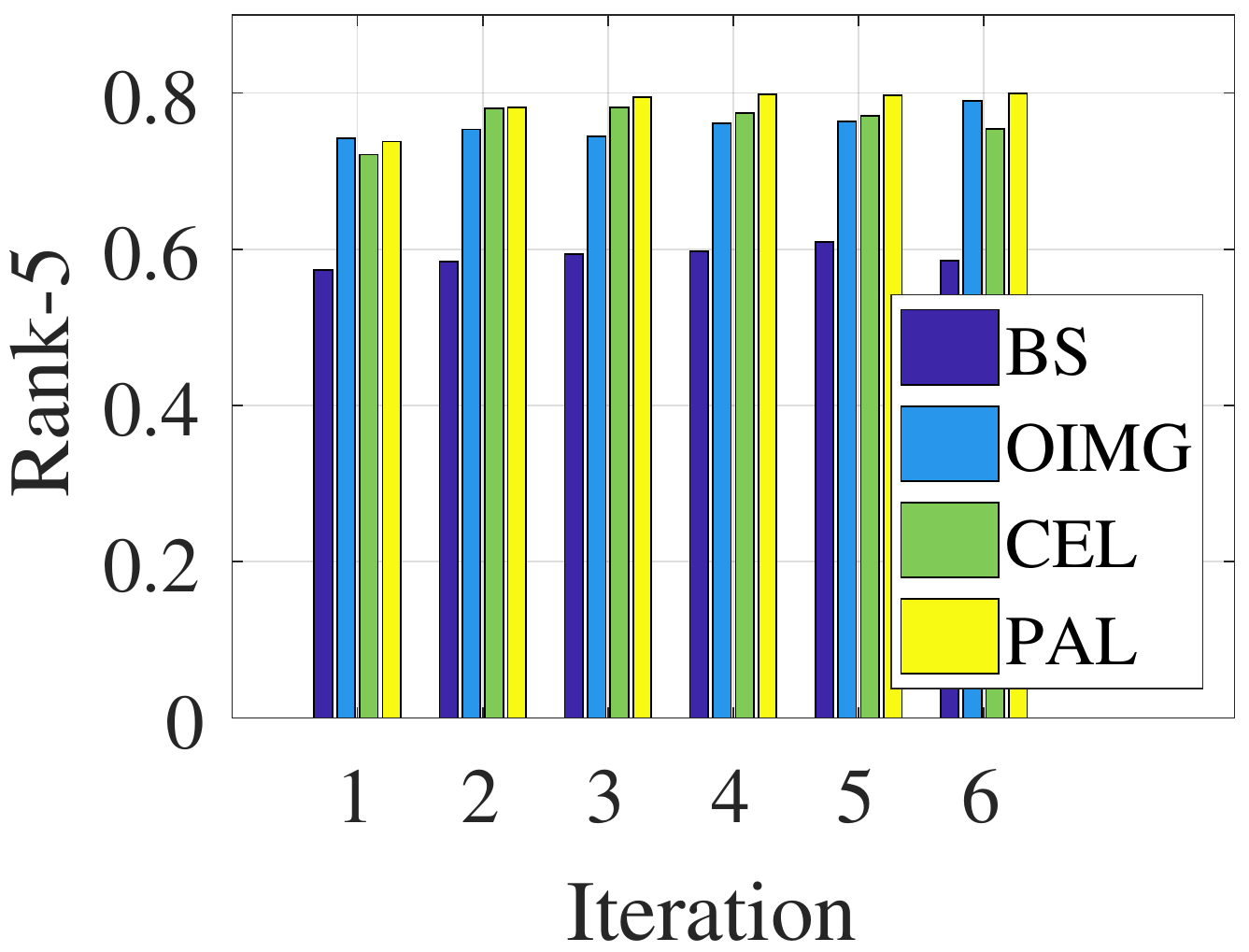}}
}
\centerline{
\subfloat[mAP]{\includegraphics[width=1.05in,height=1in]{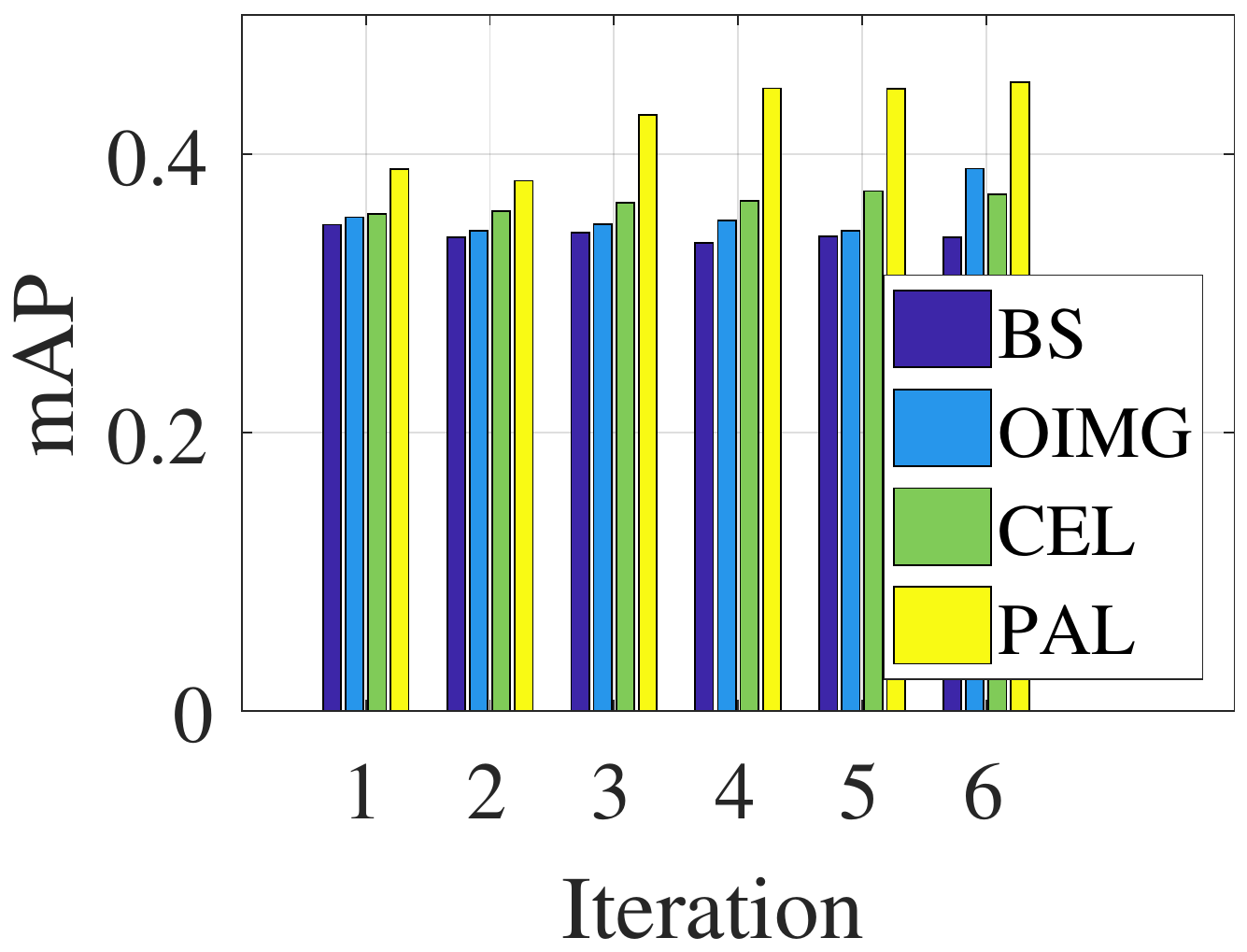}}
\subfloat[Rank-1]{\includegraphics[width=1.05in,height=1in]{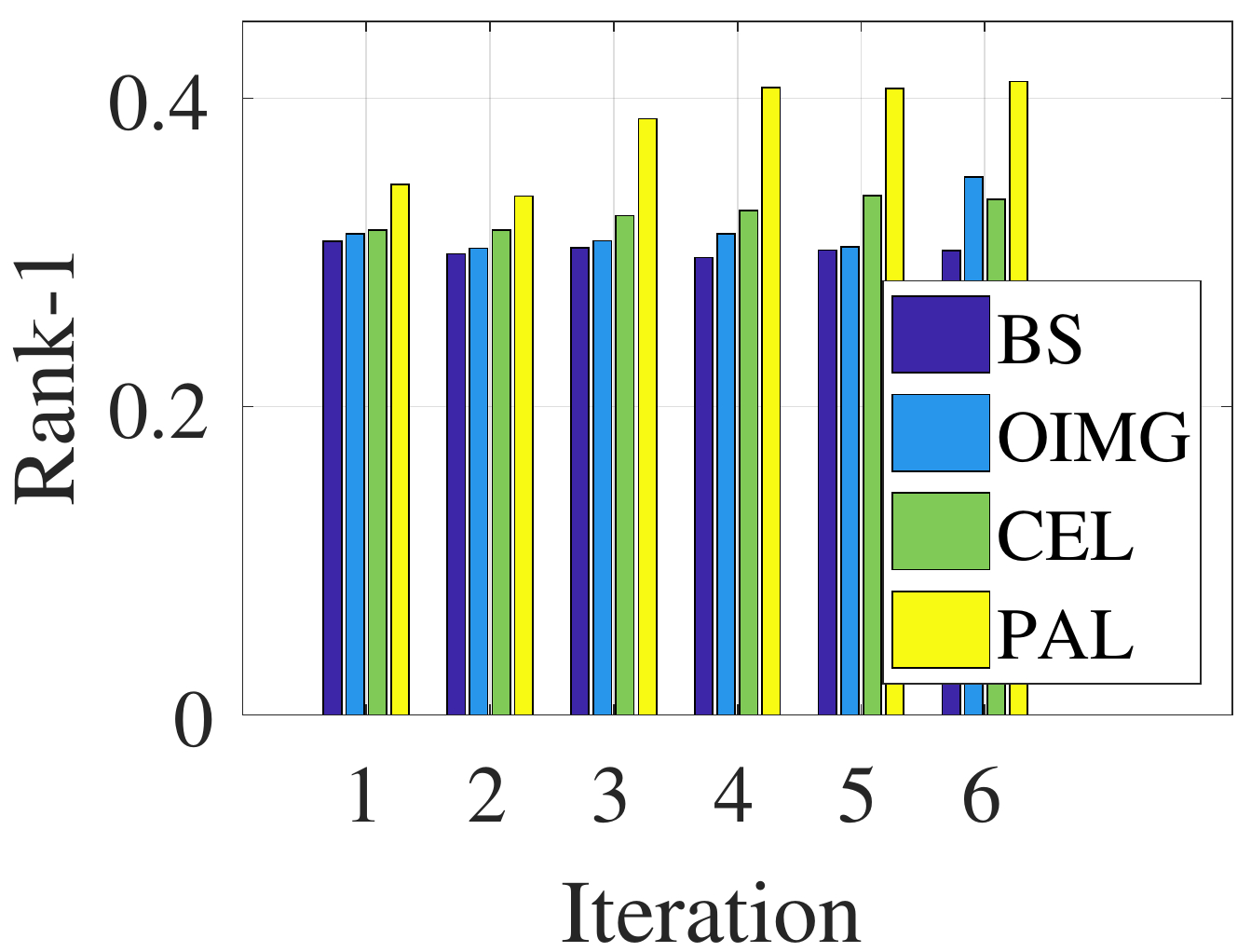}}
\subfloat[Rank-5]{\includegraphics[width=1.05in,height=1in]{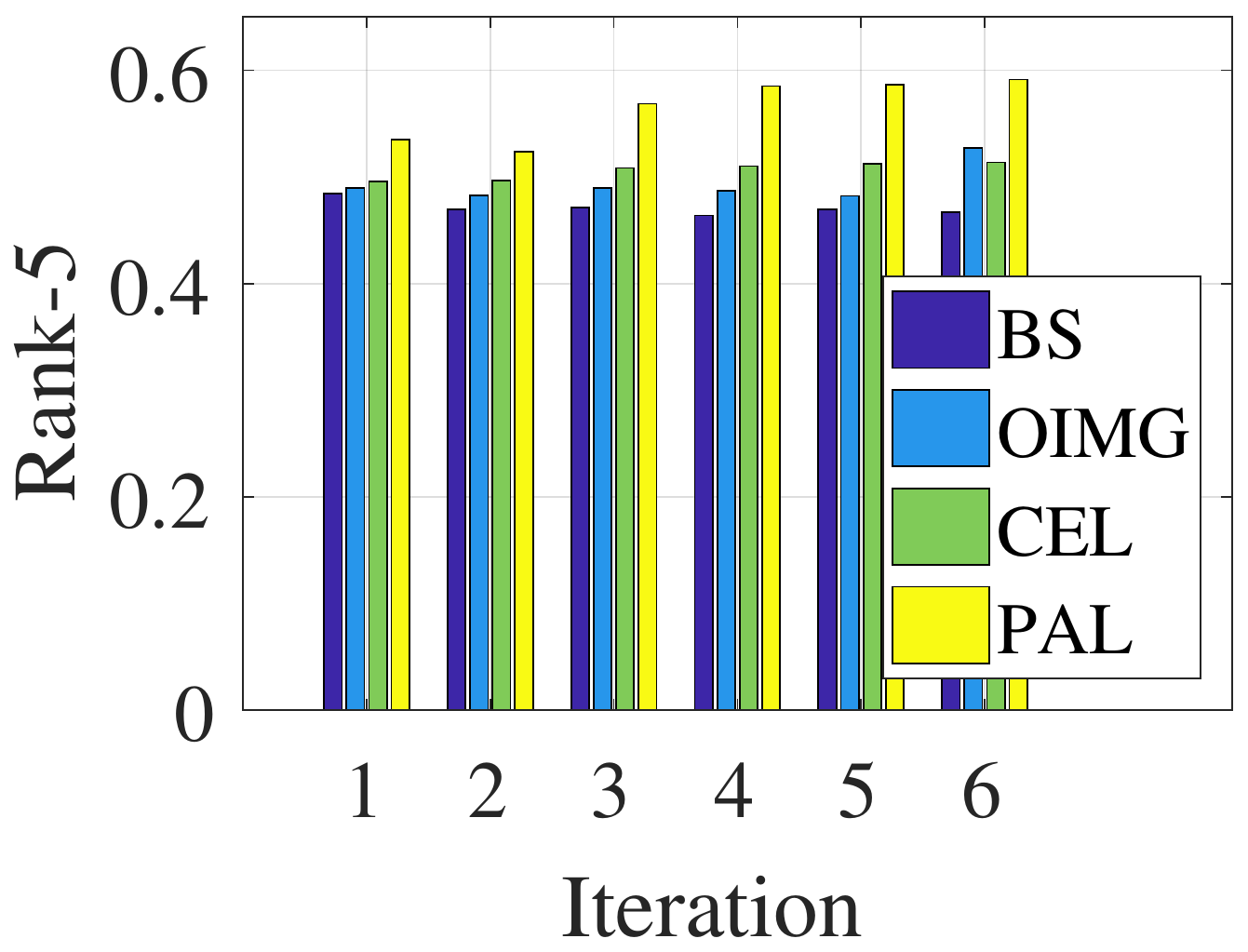}}
}
\caption{The Comparison Results. (a), (b), (c) are the mAP, Rank-1 and Rank-5 of four comparison methods on VeRi-776. (d), (e), (f) are the results of four comparison methods in every iteration on VehicleID, respectively.} \label{BarFigure}
\end{figure}

We conduct ablation studies on two major components of PAL, i.e., the data adaptation module and WLS, which are shown in Fig.\ref{BarFigure}. The settings are depicted in Table \ref{tabModels}. All of them share the similar structure with PAL. ``Generated Images'' means employing the transferred images from source domain and image of target domain to train the models, while ``Original Images'' means to utilize the original images of source domain and samples from target domain for unsupervised vehicle reID. WLS, CE represent that employ the WLS and cross-entropy loss to train reID models, respectively. Fig.\ref{BarFigure} shows that PAL achieves the best performance on two datasets, demonstrating that the data adaptation module and WLS are effective to adapt to unlabeled domain.

\begin{table}[ht]
\footnotesize
\centering
\begin{tabular}{p{0.9cm}|p{0.7cm}|p{0.7cm}|p{0.7cm}|p{0.7cm}|p{0.7cm}|p{0.7cm}}
\hline
\multirow{2}*{Iteration} & \multicolumn{3}{c|}{OIMG (\%)}  & \multicolumn{3}{c}{BS (\%)} \\
\cline{2-7} & mAP & Rank1 & Rank5 & mAP & Rank1 & Rank5\\
\hline
iter1 & 28.61 & 60.90 & 74.19    & 25.04& 57.33& 71.33\\
iter2 & 30.11 & 62.09 & 75.32    & 30.19& 58.40& 73.53\\
iter3 & 30.52 & 61.02 & 74.43    & 32.49& 59.41& 73.06\\
iter4 & 32.51 & 63.70 & 76.16    & 32.63& 59.77& 74.07\\
iter5 & 33.90 & 65.19 & 76.34    & 32.86& 60.96& 74.91\\
iter6 & 37.33 & 67.69 & 79.02    & 31.94& 58.58& 73.24\\
\hline
\end{tabular}
\caption{Performance of comparison between OIMG and BS on VeRi-776.}\label{VeRi-OIMG}
\end{table}

\begin{table}[ht]
\footnotesize
\centering
\begin{tabular}{p{0.9cm}|p{0.7cm}|p{0.7cm}|p{0.7cm}|p{0.7cm}|p{0.7cm}|p{0.7cm}}
\hline
\multirow{2}*{Iteration} & \multicolumn{3}{c|}{OIMG (\%)} & \multicolumn{3}{c}{BS (\%)} \\
\cline{2-7} & mAP & Rank1 & Rank5 & mAP & Rank1 & Rank5\\
\hline
iter1 &  35.45&  31.19 & 48.95 & 34.93& 30.71& 48.41\\
iter2 &  34.48& 30.26 & 48.25  & 34.00& 29.90& 46.98\\
iter3 &  34.94& 30.74 & 48.93  & 34.33& 30.29& 47.14\\
iter4 &  35.22& 31.20 & 48.66  & 33.62& 29.67& 46.38\\
iter5 &  34.48& 30.35 & 48.21  & 34.08& 30.15& 46.94\\
iter6 &  38.95& 34.90 & 52.69  & 34.04& 30.10& 46.94\\
\hline
\end{tabular}
\caption{Performance of comparison between OIMG and BS on VehicleID (2400).}\label{VehicleID-OIMG}
\end{table}

\paragraph{The Effectiveness of Generated Samples.}
To demonstrate the effectiveness of the generated samples, BS and CEL are compared, the results are reported in Tables \ref{VeRi-CE} and \ref{VehicleID-CE}. For CEL, we utilize CycleGAN to translate labeled images from the source domain to the target domain, and regard generated images as the ``pseudo target samples''. Then the ``pseudo target samples'' are combined with the images in the target domain to train the reID model. Both CEL and BS are trained by cross-entropy loss. According to the last iteration, compared with BS, the mAP of CEL increases by 2.09\% on VeRi-776. Besides that, it also rises to 37.12\% and 33.45\% in mAP and Rank-1 on VehicleID, demonstrating that the generated images learned the important style information from the target domain, which narrow down the domain gap.

\paragraph{The Effectiveness of WLS.}
We compare BS with OIMG to validate the effectiveness of the WLS. Tables \ref{VeRi-OIMG} and \ref{VehicleID-OIMG} show the comparisons on VeRi-776 and VehicleID, where the proposed WLS achieves better performance than cross-entropy loss. According to the last iteration, compared with the BS, the mAP and Rank-1 accuracy increased by 5.39\% and 9.11\% on VeRi-776 for OIMG, respectively. The similar conclusions hold on VehicleID, which indicates that the WLS loss has better generation ability to achieve discriminative representation during the stage of training.

\section{Conclusion}

In this paper, we propose an unsupervised vehicle reID framework, named PAL, which iteratively updates the feature learning model and estimates pseudo labels for unlabeled data for target domain adaptation. The extensive experiments of the developed algorithm has been carried out over benchmark datasets for Vehicle Re-id. It can be observed from the results that compared with other existing unsupervised methods, PAL could achieve superior performance, and even achieve better performance than some typical supervised models.

\section*{Acknowledgments}
This work was supported in part by the National Key Research and Development Program of China under grant 2018YFB0804205, by the National Natural Science Foundation of China Grant 61806035, U1936217, 61370142, 61272368, 61672365, 61732008 and 61725203, China Postdoctoral Science Foundation 3620080307, by the Dalian Science and Technology Innovation Fund 2019J11CY001, by the Fundamental Research Funds for the Central Universities Grant 3132016352, by the Liaoning Revitalization Talents Program, XLYC1908007.

\bibliographystyle{named}
\bibliography{mybibfile}

\end{document}